\documentclass{article}

\usepackage{times}
\usepackage{graphicx} 
\usepackage{float}

\usepackage{natbib}

\usepackage{algorithm}
\usepackage{algorithmic}

\usepackage[hypertexnames=false]{hyperref}

\usepackage[accepted]{icml2015}

\graphicspath{{figures/}}

\usepackage{amsmath}
\usepackage{amssymb}
\usepackage{ownstyles}

\DeclareMathOperator*{\argmax}{arg\,max}
\newcommand{\x}{\mathbf{x}}
\newcommand{\xs}{\mathbf{x^*}}
\newcommand{\xn}{\mathbf{x_0}}

\newcommand{\layer}[1]{\ensuremath{\mathsf{#1}\xspace}}
\newcommand{\unit}[2]{\ensuremath{\mathsf{#1_{#2}}\xspace}}

\newif\ifcomments

\commentstrue

\ifcomments
\newcommand{\comments}[1]{#1}
\else
\newcommand{\comments}[1]{}
\fi

\newcommand{\titl}{Understanding Neural Networks Through Deep Visualization}
\newcommand{\suptitl}{Supplementary Information for:\\\titl}
\newcommand{\suptitlrunning}{Supplementary Information for: \titl}

\icmltitlerunning{\titl}

\begin{document} 

\twocolumn[
  \icmltitle{\titl}

\icmlauthor{Jason Yosinski}{yosinski@cs.cornell.edu}
\icmladdress{Cornell University}
\icmlauthor{Jeff Clune}{jeffclune@uwyo.edu}
\icmlauthor{Anh Nguyen}{anguyen8@uwyo.edu}
\icmladdress{University of Wyoming}
\icmlauthor{Thomas Fuchs}{fuchs@caltech.edu}
\icmladdress{Jet Propulsion Laboratory, California Institute of Technology}
\icmlauthor{Hod Lipson}{hod.lipson@cornell.edu}
\icmladdress{Cornell University}

\icmlkeywords{convolutional neural networks, neural networks, visualization}

\vskip 0.3in
]

\begin{abstract}
Recent years have produced great advances in training
  large, deep neural networks (DNNs), including notable successes in training
  convolutional neural networks (convnets) to recognize natural images. However,
  our understanding of how these models work, especially what
  computations they perform at intermediate layers, has lagged
  behind. Progress in the field will be further accelerated by the development of
  better tools for visualizing and interpreting  neural nets.
  We introduce two such tools here. The first
  is a tool that visualizes the activations produced on each layer
  of a trained convnet as it processes an image or video (e.g. a live webcam stream). We have found that looking at live activations that change in response
  to user input helps build valuable intuitions about how convnets work. The second tool enables visualizing features at each
  layer of a DNN via regularized optimization in image space. Because previous versions of this idea produced less recognizable images, here
  we introduce several new regularization methods that combine to produce qualitatively clearer, more interpretable visualizations. Both tools are open source and work on a pre-trained convnet with minimal setup.
\end{abstract}

\section{Introduction}
\label{introduction}

The last several years have produced tremendous progress in training powerful, deep neural network models that are approaching and even surpassing human abilities on a variety of challenging machine learning tasks \cite{taigman-2014-CVPR-deepface:-closing-the-gap-to-human-level,schroff-2015-arXiv-facenet:-a-unified-embedding,hannun-2014-arXiv-deep-speech:-scaling}. A flagship example is training deep, convolutional neural networks (CNNs) with supervised learning to classify natural images \cite{krizhevsky2012imagenet-classification-with-deep}. That area has benefitted from the combined effects of faster computing (e.g. GPUs), better training techniques (e.g. dropout \cite{hinton2012improving-neural-networks-by-preventing}), better activation units (e.g. rectified linear units \cite{glorot-2011-AISTATS-deep-sparse-rectifier}), and larger labeled datasets  \cite{deng2009imagenet:-a-large-scale-hierarchical,lin-2014-arXiv-microsoft-coco-common}. 

While there has thus been considerable improvements in our knowledge of how to create high-performing architectures and learning algorithms, our understanding of how these large neural models operate has lagged behind. Neural networks have long been known as ``black boxes'' because it is difficult to understand exactly how any particular, trained neural network functions due to the large number of interacting, non-linear parts. Large modern neural networks are even harder to study because of their size; for example, understanding the widely-used AlexNet DNN involves making sense of the values taken by the 60 million trained network parameters. Understanding what is learned is interesting in its own right, but it is also one key way of further improving models: the intuitions provided by understanding the current generation of models should suggest ways to make them better. For example, the deconvolutional technique for visualizing the features learned by the hidden units of DNNs suggested an architectural change of smaller convolutional filters that led to state of the art performance on the ImageNet benchmark in 2013~\cite{zeiler2013visualizing-and-understanding-convolutional}.

We also note that tools that enable understanding will especially benefit the vast numbers of newcomers to deep learning, who would like to take advantage of off-the-shelf software packages --- like Theano \cite{bergstra2010theano:-a-cpu-and-gpu-math-expression}, Pylearn2 \cite{goodfellow2013pylearn2:-a-machine-learning-research}, Caffe \cite{jia2014caffe:-convolutional-architecture}, and Torch \cite{collobert-2011-torch7:-a-matlab-like-environment} --- in new domains, but who may not have any intuition for why their models work (or do not). Experts can also benefit as they iterate ideas for new models or when they are searching for good hyperparameters. We thus believe that both experts and newcomers will benefit from tools that provide intuitions about the inner workings of DNNs.  This paper provides two such tools, both of which are open source so that scientists and practitioners can integrate them with their own DNNs to better understand them. 

The first tool is software that interactively plots the activations produced on each layer of a trained DNN for user-provided images or video. Static images afford a slow, detailed investigation of a particular input, whereas video input highlights the DNNs changing responses to dynamic input.  At present, the videos are processed live from a user's computer camera, which is especially helpful because users can move different items around the field of view, occlude and combine them, and perform other manipulations to actively learn how different features in the network respond.

\begin{figure*}[!th]
\begin{center}
  \centerline{\includegraphics[width=1\linewidth]{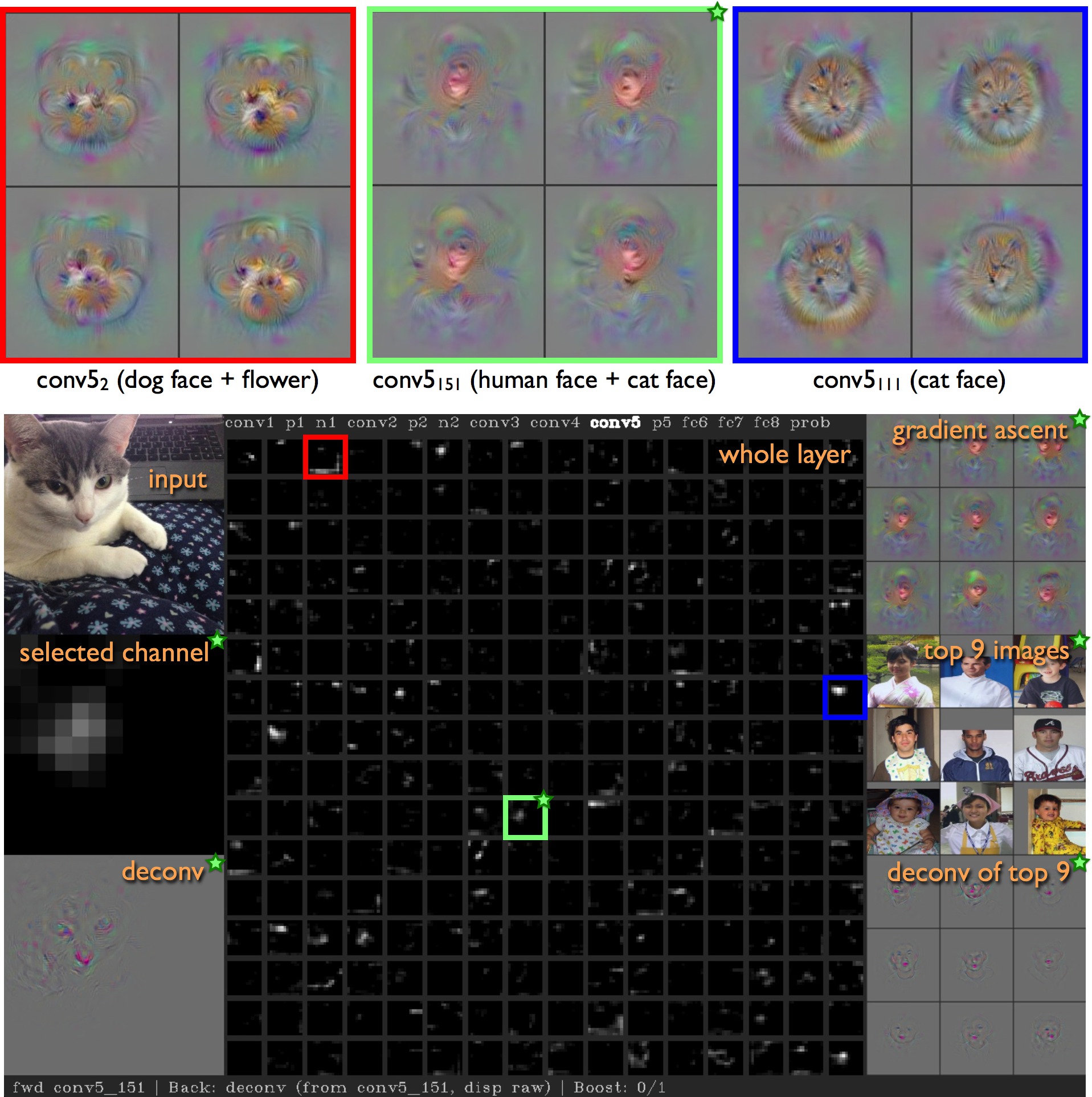}}
  \vskip -.1in
\caption{The \textbf{bottom} shows a screenshot from the interactive visualization software. The webcam \emph{input} is shown, along with the \emph{whole layer} of \layer{conv5} activations. The \emph{selected channel} pane shows an enlarged version of the 13x13 \unit{conv5}{151} channel activations. Below it, the \emph{deconv} starting at the selected channel is shown. On the right, three selections of nine images are shown: synthetic images produced using the regularized \emph{gradient ascent} methods described in \secref{optimization}, the \emph{top 9 image} patches from the training set
  (the images from the training set that caused the highest activations for the selected channel),
  and the \emph{deconv of the those top 9} images. All areas highlighted with a green star relate to the particular selected channel, here \unit{conv5}{151}; when the selection changes, these panels update.
The \textbf{top} depicts enlarged numerical optimization results for this and other channels. \unit{conv5}{2} is a channel that responds most strongly to dog faces (as evidenced by the top nine images, which are not shown due to space constraints), but it also responds to flowers on the blanket on the bottom and half way up the right side of the image (as seen in the inset red highlight). This response to flowers can be partially seen in the optimized images but would be missed in an analysis focusing only on the top nine images and their deconv versions, which contain no flowers.
\unit{conv5}{151} detects different types of faces. The top nine images are all of human faces, but here we see it responds also to the cat's face (and in \figref{demo_face} a lion's face). Finally, \unit{conv5}{111} activates strongly for the cat's face, the optimized images show catlike fur and ears, and the top nine images (not shown here) are also all of cats. For this image, the \layer{softmax} output layer top two predictions are ``Egyptian Cat'' and ``Computer Keyboard.'' All figures in this paper are best viewed digitally, in color, significantly zoomed in.
}
\figlabel{demo_layers}
\end{center}
\vskip -0.2in
\end{figure*}



The second tool we introduce enables better visualization of the learned features computed by individual neurons at every layer of a DNN. Seeing what features have been learned is important both to understand how current DNNs work and to fuel intuitions for how to improve them. 

Attempting to understand what computations are performed at each layer in DNNs is an increasingly popular direction of research. One approach is to study each layer as a group and investigate the type of computation performed by the set of neurons on a layer as a whole~\cite{yosinski-2014-NIPS-how-transferable-are-features-in-deep,mahendran-2014-arXiv-understanding-deep-image}. This approach is informative because the neurons in a layer interact with each other to pass information to higher layers, and thus each neuron's contribution to the entire function performed by the DNN depends on that neuron's context in the layer. 

Another approach is to try to interpret the function computed by each individual neuron. Past studies in this vein roughly divide into two different camps: \emph{dataset-centric} and \emph{network-centric}. The former requires both a trained DNN and running data through that network; the latter requires only the trained network itself. One dataset-centric approach is to display images from the training or test set that cause high or low activations for individual units. Another is the deconvolution method of Zeiler \& Fergus~\yrcite{zeiler2013visualizing-and-understanding-convolutional}, which highlights the portions of a particular image that are responsible for the firing of each neural unit.

Network-centric approaches investigate a network directly without any data from a dataset. For example, Erhan et al.~\yrcite{erhan2009visualizing-higher-layer-features} synthesized images that cause high activations for particular units. Starting with some initial input $\x = \xn$, the activation $a_i(\x)$ caused at some unit $i$ by this input is computed, and then steps are taken in input space
along the gradient $\partial a_i(\x) /\partial \x$ to synthesize inputs that cause higher and higher activations of unit $i$, eventually terminating at some $\xs$ which is deemed to be a preferred input stimulus for the unit in question. In the case where the input space is an image, $\xs$ can be displayed directly for interpretation. Others have followed suit, using the gradient to find images that cause higher activations \cite{simonyan2013deep-inside-convolutional,nguyen-2014-arXiv-deep-neural-networks} or lower activations \cite{szegedy2013intriguing-properties-of-neural} for output units.

These gradient-based approaches are attractive in their simplicity, but the optimization process tends to produce images that do not greatly resemble natural images. Instead, they are composed of a collection of ``hacks'' that happen to cause high (or low) activations: extreme pixel values, structured high frequency patterns, and copies of common motifs without global structure~\cite{simonyan2013deep-inside-convolutional,nguyen-2014-arXiv-deep-neural-networks,szegedy2013intriguing-properties-of-neural,goodfellow-2014-arXiv-explaining-and-harnessing-adversarial}. The fact that activations may be effected by such hacks is better understood thanks to several recent studies. Specifically, it has been shown that such hacks may be applied to correctly classified images to cause them to be misclassified even via imperceptibly small changes \cite{szegedy2013intriguing-properties-of-neural}, that such hacks can be found even without the gradient information to produce unrecognizable ``fooling examples'' \cite{nguyen-2014-arXiv-deep-neural-networks}, and that the abundance of non-natural looking images that cause extreme activations can be explained by the locally linear behavior of neural nets 
\cite{goodfellow-2014-arXiv-explaining-and-harnessing-adversarial}.

With such strong evidence that optimizing images to cause high activations produces unrecognizable images, is there any hope of using such methods to obtain useful visualizations? It turns out there is, if one is able to appropriately regularize the optimization. Simonyan et al.~\yrcite{simonyan2013deep-inside-convolutional} showed that slightly discernible images for the final layers of a convnet could be produced with $L_2$-regularization. Mahendran and Vedaldi \yrcite{mahendran-2014-arXiv-understanding-deep-image} also showed the importance of incorporating natural-image priors in the optimization process when producing images that mimic an entire-layer's firing pattern produced by a specific input image. We build on these works and contribute three additional forms of regularization that, when combined, produce more recognizable, optimization-based samples than previous methods. Because the optimization is stochastic, by starting at different random initial images, we can produce a set of optimized images whose variance provides information about the invariances learned by the unit.

To summarize, this paper makes the following two contributions:

\begin{enumerate}

\item We describe and release a software tool that provides a live, interactive visualization of every neuron in a trained convnet as it responds to a user-provided image or video. The tool displays forward activation values, preferred stimuli via gradient ascent, top images for each unit from the training set, deconv highlighting \cite{zeiler2013visualizing-and-understanding-convolutional} of top images, and backward diffs computed via backprop or deconv starting from arbitrary units. The combined effect of these complementary visualizations promotes a greater understanding of what a neuron computes than any single method on its own. We also describe a few insights we have gained from using this tool.
(\secref{demo}).


\item We extend past efforts to visualize preferred activation patterns in input space by adding several new types of regularization, which produce what we believe are the most interpretable images for large convnets so far (\secref{optimization}).

\end{enumerate}

Both of our tools are released as open source and are available at
\url{http://yosinski.com/deepvis}. While the tools could be adapted to integrate with any DNN software framework, they work out of the box with 
the popular Caffe DNN software package \cite{jia2014caffe:-convolutional-architecture}.
Users may run visualizations with their own Caffe DNN or our pre-trained DNN, which comes with pre-computed images optimized to activate each neuron in this trained network. Our pre-trained network is nearly identical to the ``AlexNet'' architecture \cite{krizhevsky2012imagenet-classification-with-deep}, but with local reponse normalization layers after pooling layers following \citep{jia2014caffe:-convolutional-architecture}. It was trained with the Caffe framework on the ImageNet 2012 dataset \cite{deng2009imagenet:-a-large-scale-hierarchical}.

\section{Visualizing Live Convnet Activations}
\seclabel{demo}

Our first visualization method is straightforward:  plotting the activation values for the neurons in each layer of a convnet in response to an image or video. In fully connected neural networks, the order of the units is irrelevant, so plots of these vectors are not spatially informative. However, in convolutional networks, filters are applied in a way that respects the underlying geometry of the input; in the case of 2D images, filters are applied in a 2D convolution over the two spatial dimensions of the image. This convolution produces activations on subsequent layers that are, for each channel, also arranged spatially.

\figref{demo_layers} shows examples of this type of plot for the \layer{conv5} layer.
The \layer{conv5} layer has size 256$\times$13$\times$13, which we depict as 256 separate 13$\times$13 grayscale images. Each of the 256 small images contains activations in the same spatial $x$-$y$ spatial layout as the input data, and the 256 images are simply and arbitrarily tiled into a 16$\times$16 grid in row-major order.
\figref{demo_face} shows a zoomed in view of one particular channel, \unit{conv5}{151}, that responds to human and animal faces. All layers can be viewed in the software tool, including pooling and normalization layers. Visualizing these layers provides intuitions about their effects and functions.

\begin{figure}[!h]
\vskip 0.2in
\begin{center}
  \includegraphics[width=.5\linewidth]{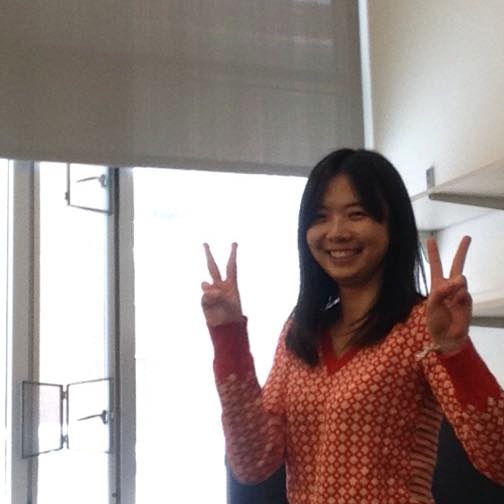}\includegraphics[width=.5\linewidth]{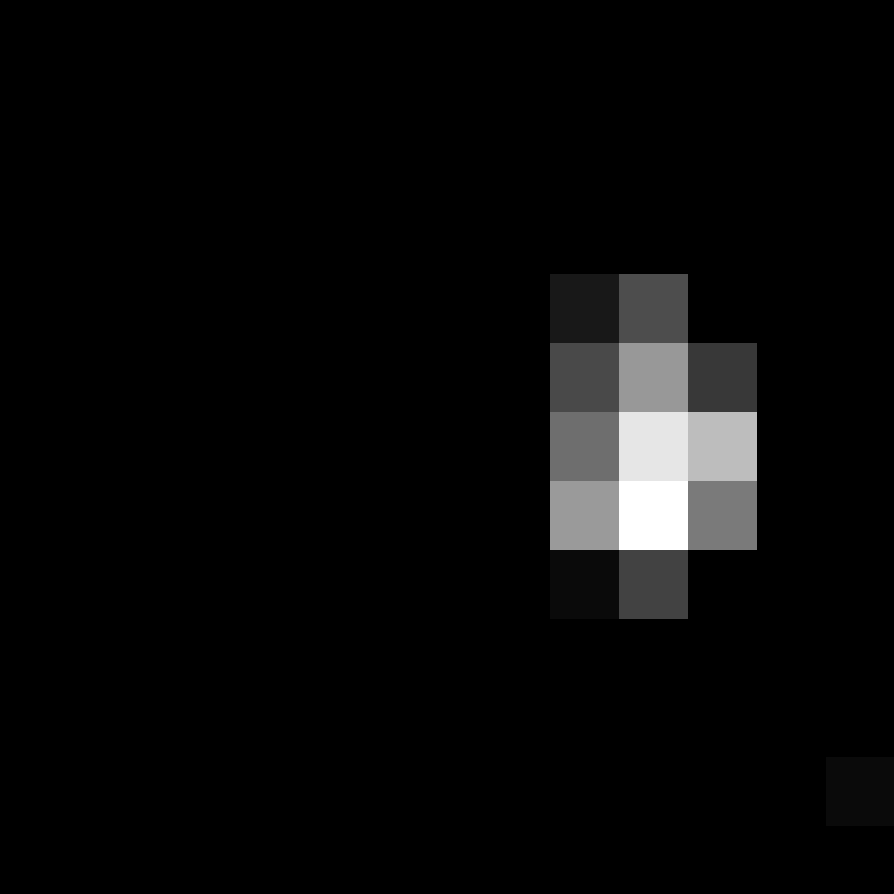} \\
  \vspace{.5ex}
  \includegraphics[width=.5\linewidth]{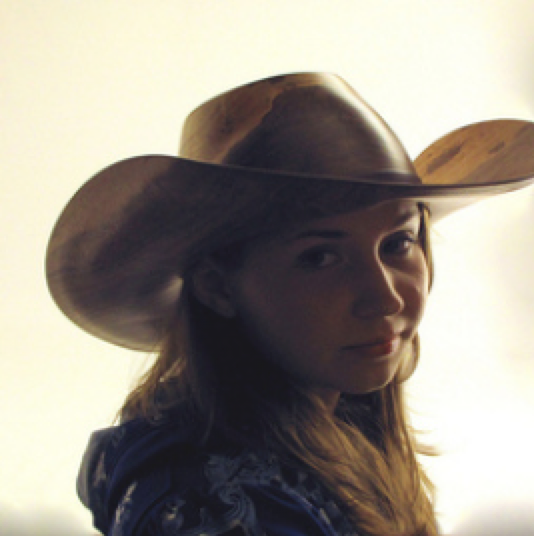}\includegraphics[width=.5\linewidth]{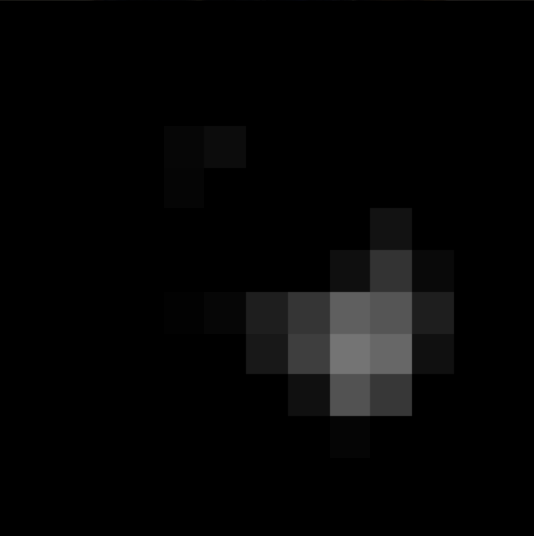} \\
  \vspace{.5ex}
  \includegraphics[width=.5\linewidth]{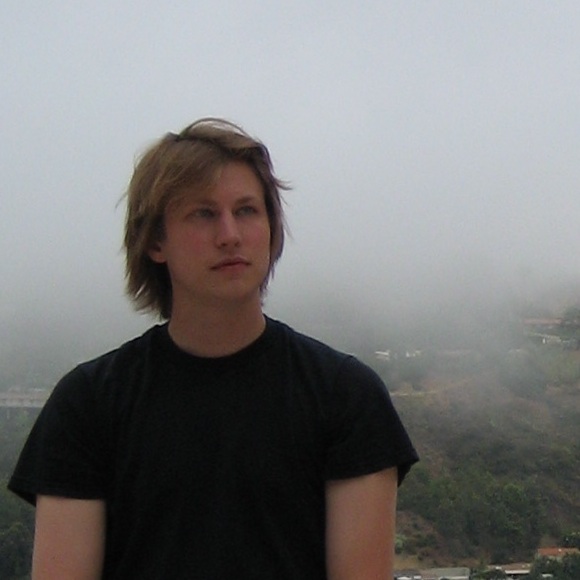}\includegraphics[width=.5\linewidth]{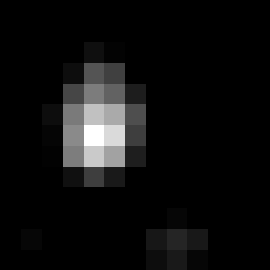} \\
  \vspace{.5ex}
  \includegraphics[width=.5\linewidth]{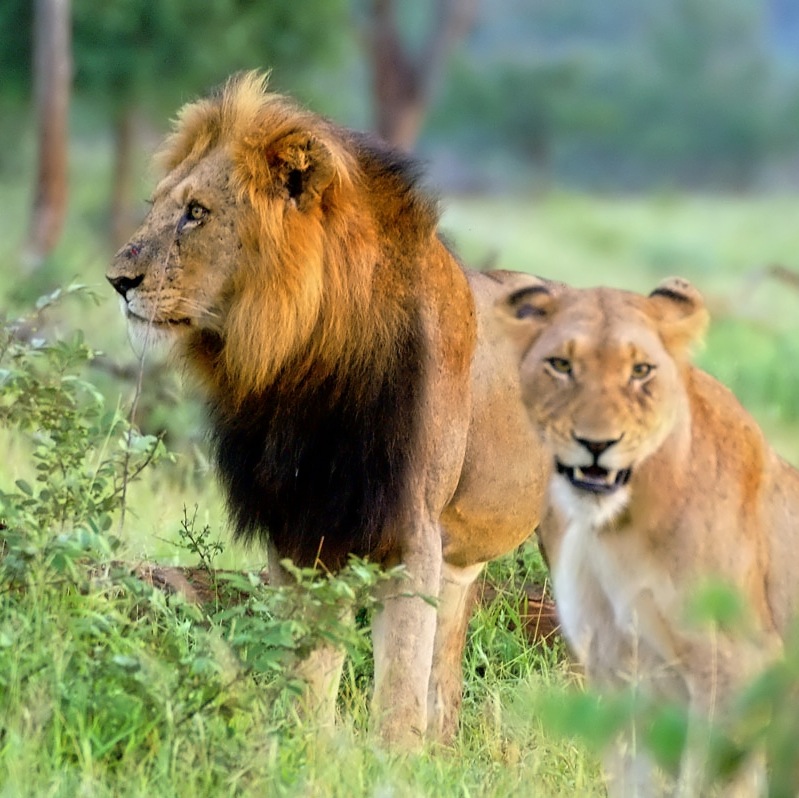}\includegraphics[width=.5\linewidth]{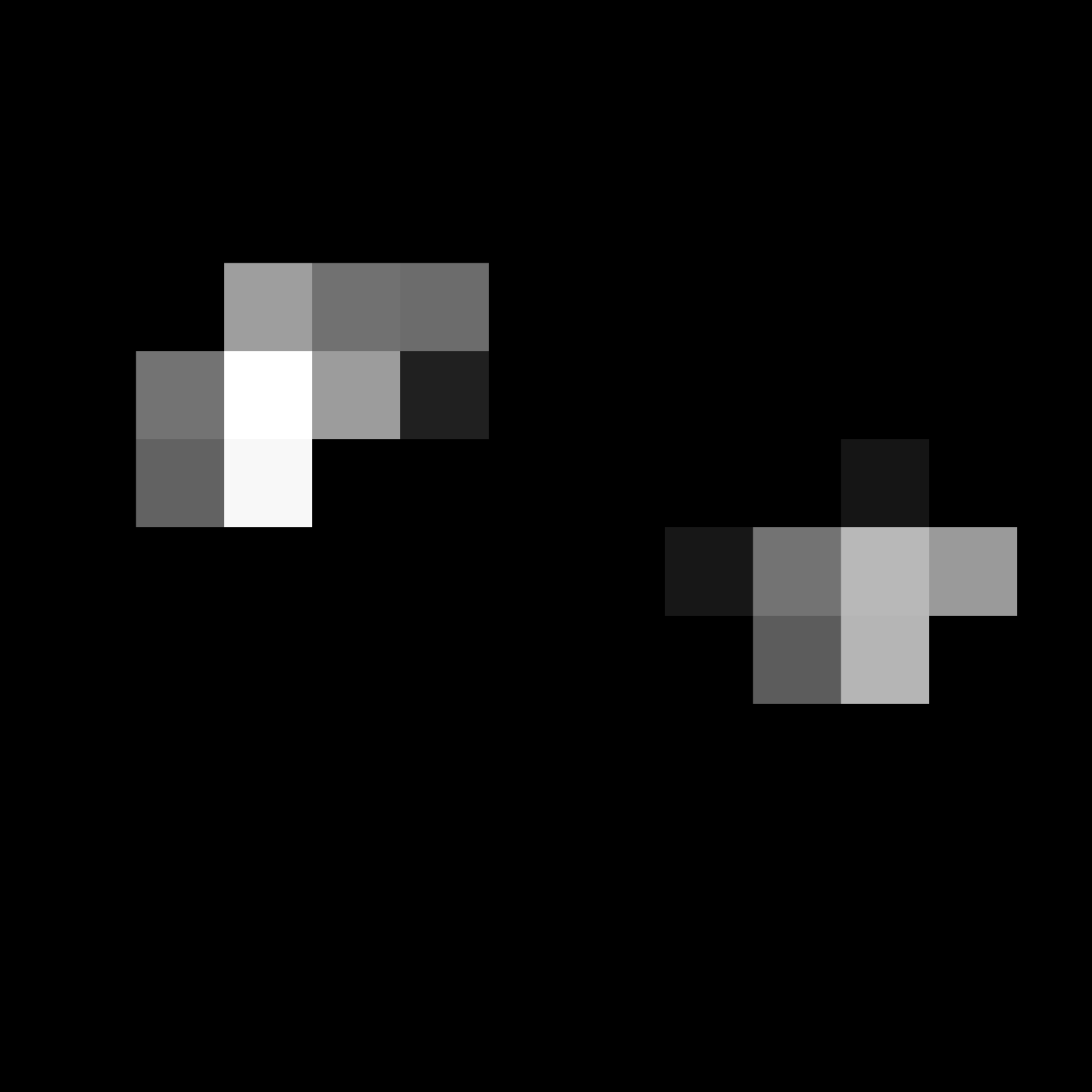}
  \caption{A view of the 13$\times$13 activations of the 151\textsuperscript{st} channel on the \layer{conv5} layer of a deep neural network trained on ImageNet, a dataset that does not contain a face class, but does contain many images with faces. The channel  responds to human and animal faces and is robust to changes in scale, pose, lighting, and context, which can be discerned by a user by actively changing the scene in front of a webcam or by loading static images (e.g. of the lions) and seeing the corresponding response of the unit.
    Photo of lions via Flickr user arnolouise, licensed under CC BY-NC-SA 2.0.
  }
  \figlabel{demo_face}
\end{center}
\vskip -0.2in
\end{figure}

Although this visualization is simple to implement, we find it informative because
all data flowing through the network can be visualized. There is nothing mysterious happening behind the scenes. Because this convnet contains only a single path from input to output, every layer is a bottleneck through which all information must pass en-route to a classification decision. The layer sizes are all small enough that any one layer can easily fit on a computer screen.\footnote{The layer with the most activations is \layer{conv1} which, when tiled, is only 550x550 before adding padding.} So far, we have gleaned several surprising intuitions from using the tool:


\begin{itemize}


\item One of the most interesting conclusions so far has been that representations on some layers seem to be surprisingly local. Instead of finding distributed representations on all layers, we see, for example, detectors for text, flowers, fruit, and faces on \layer{conv4} and \layer{conv5}. These conclusions can be drawn either from the live visualization or the optimized images (or, best, by using both in concert) and suggest several directions for future research (discussed in \secref{conclusion}).

\item When using direct file input to classify photos from Flickr or Google Images, classifications are often correct and highly confident (softmax probability for correct class near 1). On the other hand, when using input from a webcam, predictions often cannot be correct because no items from the training set are shown in the image. The training set's 1000 classes, though numerous, do not cover most common household objects. Thus, when shown a typical webcam view of a person with no ImageNet classes present, the output has no single high probability, as is expected. Surprisingly, however, this probability vector is noisy and varies significantly in response to tiny changes in the input, often changing merely in response to the noise from the webcam. We might have instead expected unchanging and low confidence predictions for a given scene when no object the network has been trained to classify is present. Plotting the fully connected layers (\layer{fc6} and \layer{fc7}) also reveals a similar sensitivity to small input changes. 

\item Although the last three layers are sensitive to small input changes, much of the lower layer computation is more robust. For example, when visualizing the \layer{conv5} layer, one can find many invariant detectors for faces, shoulders, text, etc. by moving oneself or objects in front of the camera. Even though the 1000 classes contain no explicitly labeled faces or text, the network learns to identify these concepts simply because they represent useful partial information for making a later classification decision. One face detector, denoted \unit{conv5}{151} (channel number 151 on \layer{conv5}), is shown in \figref{demo_face} activating for human and lion faces and in \figref{demo_layers} activating for a cat face. Zhou et al.~\yrcite{zhou-2014-arXiv-object-detectors-emerge} recently observed a similar effect where convnets trained only to recognize different scene types --- playgrounds, restaurant patios, living rooms, etc. --- learn object detectors (e.g. for chairs, books, and sofas) on intermediate layers.

\end{itemize}

The reader is encouraged to try this visualization tool out for him or herself. The code, together with pre-trained models and images synthesized by gradient ascent, can be downloaded at
\url{http://yosinski.com/deepvis}.

\section{Visualizing via Regularized Optimization}
\seclabel{optimization}

The second contribution of this work is introducing several regularization methods to bias images found via optimization toward more visually interpretable examples. While each of these regularization methods helps on its own, in combination they are even more effective. We found useful combinations via a random hyperparameter search, as discussed below.

\seclabel{reg_methods}

Formally, consider an image $\x \in \mathbb{R} ^ {C \times H \times W}$, where\ $C = 3$ color channels and the height ($H$) and width ($W$) are both 227 pixels. When this image is presented to a neural network, it causes an activation $a_i(\x)$ for some unit $i$, where for simplicity $i$ is an index that runs over all units on all layers. We also define a parameterized regularization function $R_\theta(\x)$ that penalizes images in various ways.

Our network was trained on ImageNet by first subtracting the per-pixel mean of examples in ImageNet before inputting training examples to the network. Thus, the direct input to the network, $\x$, can be thought of as a zero-centered input.  We may pose the optimization problem as finding an image $\xs$ where

\be
\xs = \argmax_\x(a_i(\x) - R_{\theta}(\x))
\ee

In practice, we use a slightly different formulation. Because we search for $\xs$ by starting at some $\xn$ and taking gradient steps, we instead define the regularization via an operator $r_\theta(\cdot)$ that maps $\x$ to a slightly more regularized version of itself. This latter definition is strictly more expressive, allowing regularization operators $r_\theta$ that are not the gradient of any $R_\theta$.
This method is easy to implement within a gradient descent framework by simply alternating between taking a step toward the gradient of $a_i(\x)$ and taking a step in the direction given by $r_\theta$. With a gradient descent step size of $\eta$, a single step in this process applies the update:

\be
\x \leftarrow r_\theta\left(\x + \eta\frac{\partial a_i}{\partial \x}\right) \\
\ee

We investigated the following four regularizations. All are designed to overcome different pathologies commonly encountered by gradient descent without regularization.

{\bf $L_2$ decay}: A common regularization, $L_2$ decay penalizes large values and is implemented as $r_\theta(\x) =(1 - \theta_{\mathrm{decay}})\cdot\x$. $L_2$ decay tends to prevent a small number of extreme pixel values from dominating the example image. Such extreme single-pixel values neither occur naturally with great frequency nor are useful for visualization. $L_2$ decay was also used by Simonyan et al.~\yrcite{simonyan2013deep-inside-convolutional}.

{\bf Gaussian blur}: Producing images via gradient ascent tends to produce examples with high frequency information (see Supplementary \secref{high_freq} for a possible reason). While these images cause high activations, they are neither realistic nor interpretable \cite{nguyen-2014-arXiv-deep-neural-networks}. A useful regularization is thus to penalize high frequency information. We implement this as a Gaussian blur step $r_\theta(\x) = \mathrm{GaussianBlur}(\x, \theta_{\mathrm{b\_width}})$. Convolving with a blur kernel is more computationally expensive than the other regularization methods, so we added another hyperparameter $\theta_{\mathrm{b\_every}}$ to allow, for example, blurring every several optimization steps instead of every step. Blurring an image multiple times with a small width Gaussian kernel is equivalent to blurring once with a larger width kernel, and the effect will be similar even if the image changes slightly during the optimization process. This technique thus lowers computational costs without limiting the expressiveness of the regularization. Mahendran \& Vedaldi~\yrcite{mahendran-2014-arXiv-understanding-deep-image} used a penalty with a similar effect to blurring, called \emph{total variation}, in their work reconstructing images from layer codes.

{\bf Clipping pixels with small norm}: The first two regularizations suppress high amplitude and high frequency information, so after applying both, we are left with an $\xs$ that contains somewhat small, somewhat smooth values. However, $\xs$ will still tend to contain non-zero pixel values everywhere. Even if some pixels in $\xs$ show the primary object or type of input causing the unit under consideration to activate, the gradient with respect to all other pixels in $\xs$ will still generally be non-zero, so these pixels will also shift to show some pattern as well, contributing in whatever small way they can to ultimately raise the chosen unit's activation. We wish to bias the search away from such behavior and instead show only the main object, letting other regions be exactly zero if they are not needed. We implement this bias using an $r_\theta(\x)$ that computes the norm of each pixel (over red, green, and blue channels) and then sets any pixels with small norm to zero. The threshold for the norm, $\theta_{\mathrm{n\_pct}}$, is specified as a percentile of all pixel norms in $\x$.

{\bf Clipping pixels with small contribution}: Instead of clipping pixels with small norms, we can try something slightly smarter and clip pixels with small \emph{contributions} to the activation. One way of computing a pixel's contribution to an activation is to measure how much the activation increases or decreases when the pixel is set to zero; that is, to compute the contribution as $|a_i(\x) - a_i(\x_{-j})|$, where $\x_{-j}$ is $\x$ but with the $j^{th}$ pixel set to zero. This approach is straightforward but prohibitively slow, requiring a forward pass for every pixel. Instead, we approximate this process by linearizing $a_i(\x)$ around $\x$, in which case the contribution of each dimension of $\x$ can be estimated as the elementwise product of $\x$ and the gradient. We then sum over all three channels and take the absolute value, computing  $\left|\sum_c \x \circ \nabla_\x a_i(\x)\right|$. We use the absolute value to find pixels with small contribution in either direction, positive or negative. While we could choose to keep the pixel transitions where setting the pixel to zero would result in a large activation increase, these shifts are already handled by gradient ascent, and here we prefer to clip only the pixels that are deemed not to matter, not to take large gradient steps outside the region where the linear approximation is most valid. We define this $r_\theta(\x)$ as the operation that sets pixels with contribution under the $\theta_{\mathrm{c\_pct}}$ percentile to zero.



\seclabel{results}

\begin{figure}[!th]
\begin{center}
\centerline{\includegraphics[width=1.0\linewidth]{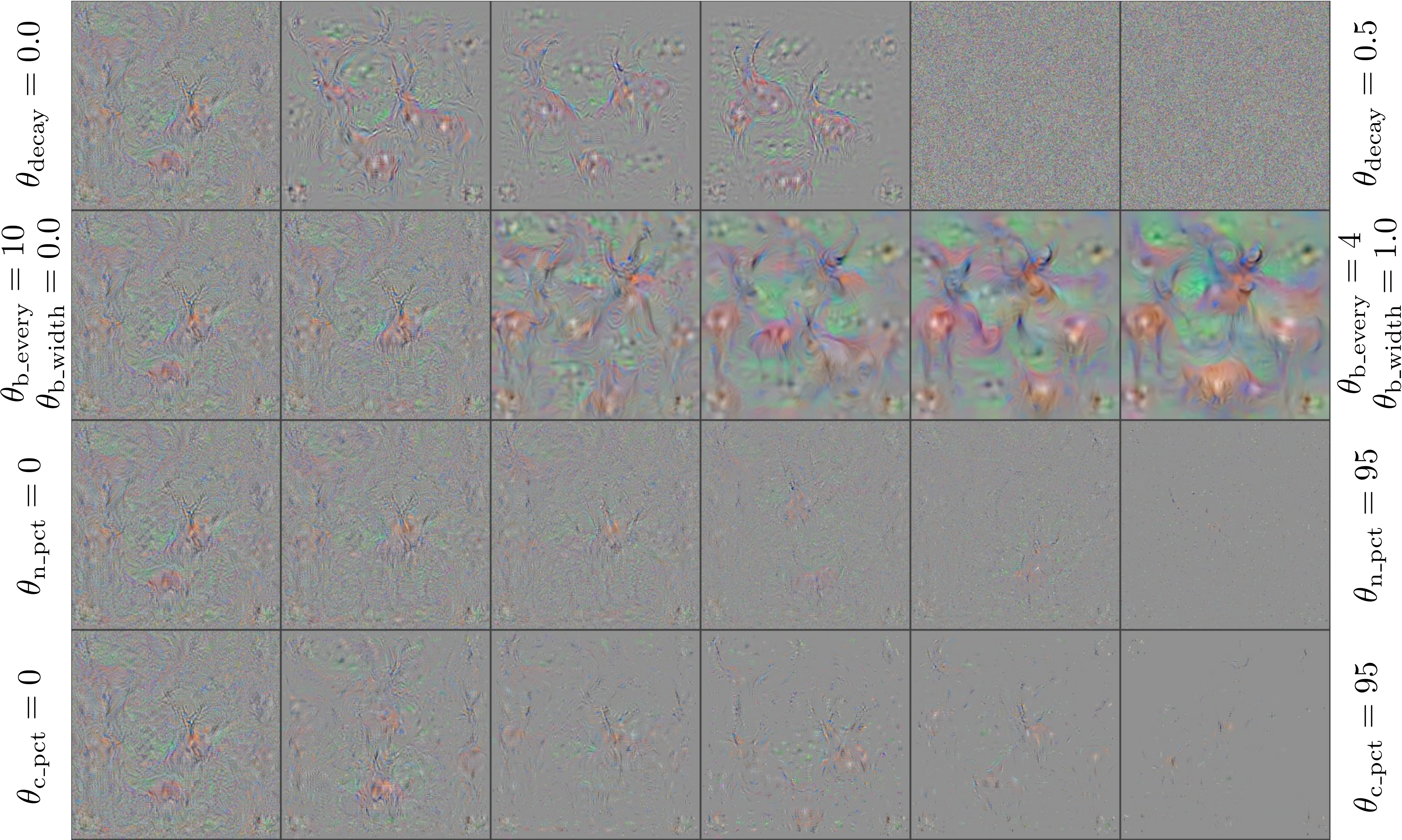}}
\caption{The effects of each regularization method from \secref{optimization} when used individually. Each of the four rows shows a linear sweep in hyperparameter space from no regularization (left) to strong regularization (right).
  When applied too strongly, some regularizations cause the optimization to fail (e.g. $L_2$ decay, top row) or the images to be less interpretable (small norm and small contribution clipping, bottom two rows). For this reason, a random hyperparameter search was useful for finding joint hyperparameter settings that worked well together (see \figref{vis_fc8}).
Best viewed electronically, zoomed in.
}
\figlabel{regularization_sweep}
\end{center}
\end{figure}

If the above regularization methods are applied individually, they are
somewhat effective at producing more interpretable images; \figref{regularization_sweep} shows the effects of each individual hyperparameter.
However, preliminary experiments uncovered that their combined
effect produces better visualizations. To pick a reasonable set of
hyperparameters for all methods at once, we ran a random
hyperparameter search of 300 possible combinations and settled on four
that complement each other well. The four selected combinations are
listed in \tabref{paramTable} and optimized images using each are shown for the ``Gorilla'' class output unit in \figref{vis_fc8}. Of the four, some show high
frequency information, others low frequency; some contain dense
pixel data, and others contain only sparse outlines of important
regions.
We found the version in the lower-left quadrant to be the best single set of hyperparameters, but often greater intuition can
be gleaned by considering all four at once.
\figref{vis_all} shows the optimization results computed for a selection of units on all layers. A single image for every filter of all five convolutional layers is shown in Supplementary \figref{layer_montages}. Nine images for each filter of all layers, including each of the 1000 ImageNet output classes, can be viewed at \url{http://yosinski.com/deepvis}.

\begin{table}[t]
\caption{Four hyperparameter combinations that produce different styles of recognizable images. We identified these four after reviewing images produced by 300 randomly selected hyperparameter combinations. From top to bottom, they are the hyperparameter combinations that produced the top-left, top-right, bottom-left, and bottom-right Gorilla class visualizations, respectively, in \figref{vis_fc8}. The third row hyperparameters produced most of the visualizations for the other classes in \figref{vis_fc8}, and all of those in \figref{vis_all}.
}
\tablabel{boost}
\begin{center}
\begin{tabular}{|c|c|c|c|c|}
  \hline
  $\theta_{\mathrm{decay}}$ & $\theta_{\mathrm{b\_width}}$ & $\theta_{\mathrm{b\_every}}$ & $\theta_{\mathrm{n\_pct}}$ & $\theta_{\mathrm{c\_pct}}$  \\
  \hline
    0    &   0.5   &     4    &    50    &    0 \\
    0.3  &    0    &     0    &    20    &    0 \\
  0.0001 &   1.0   &     4    &     0    &    0 \\
    0    &   0.5   &     4    &     0    &   90 \\
\hline
\end{tabular}
\end{center}
\tablabel{paramTable}
\end{table}

\begin{figure*}[!ht]
\vskip -0.2in
\begin{center}
\centerline{\includegraphics[width=.95\linewidth]{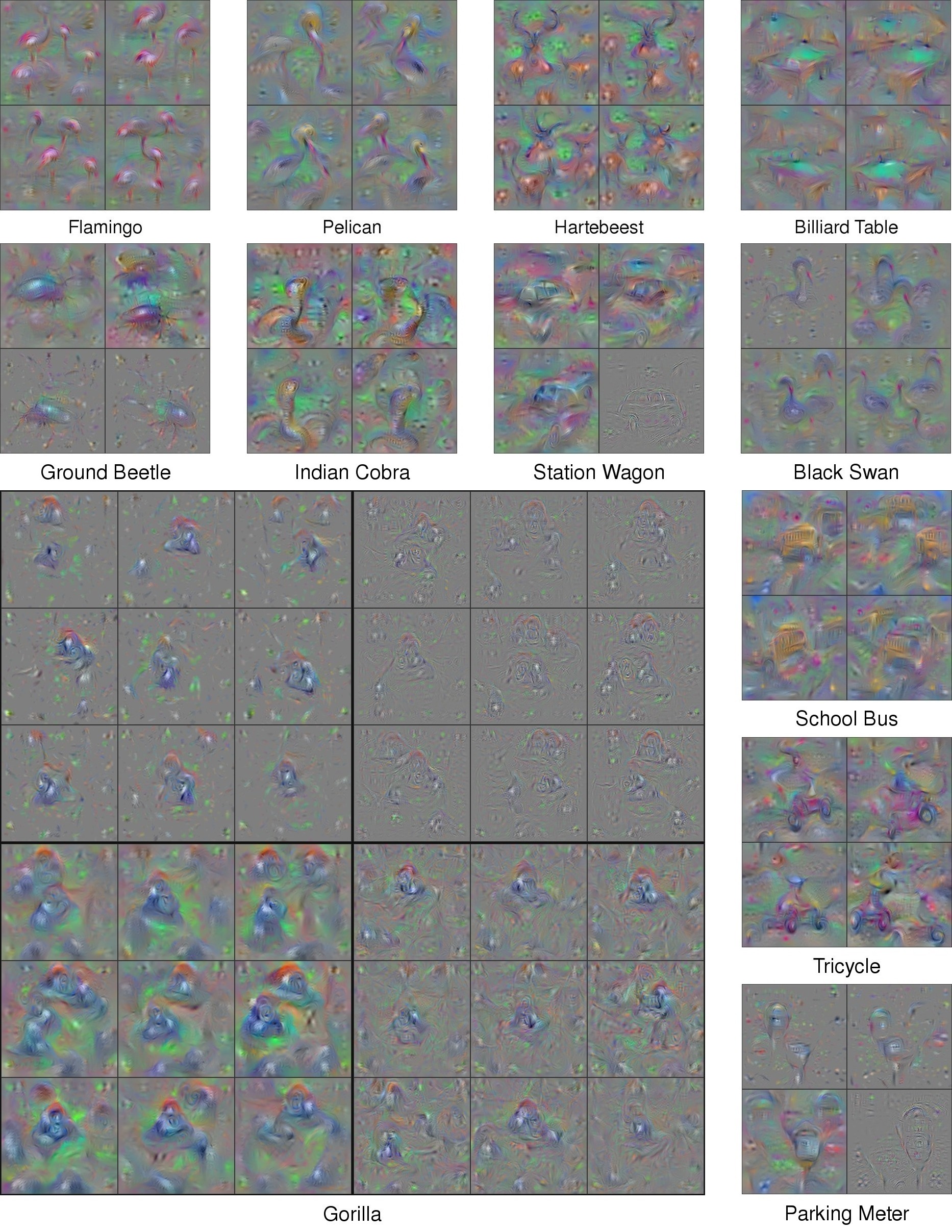}}
\vskip -0.2in
\caption{Visualizations of the preferred inputs for different class units on layer \layer{fc8}, the 1000-dimensional output of the network just before the final softmax. In the lower left are 9 visualizations each (in 3$\times$3 grids) for four different sets of regularization hyperparameters for the Gorilla class (\tabref{paramTable}). For all other classes, we have selected four interpretable visualizations produced by our regularized optimization method. We chose the four combinations of regularization hyperparameters by performing a random hyperparameter search and selecting combinations that complement each other. For example, the lower left quadrant tends to show lower frequency patterns, the upper right shows high frequency patterns, and the upper left shows a sparse set of important regions.  Often greater intuition can be gleaned by considering all four at once. In nearly every case, we have found that one can guess what class a neuron represents by viewing sets of these optimized, preferred images. Best viewed electronically, zoomed in.
}
\figlabel{vis_fc8}
\end{center}
\vskip -0.9in
\end{figure*}

\begin{figure*}[!ht]
\vskip 0.2in
\begin{center}
\centerline{\includegraphics[width=1\linewidth]{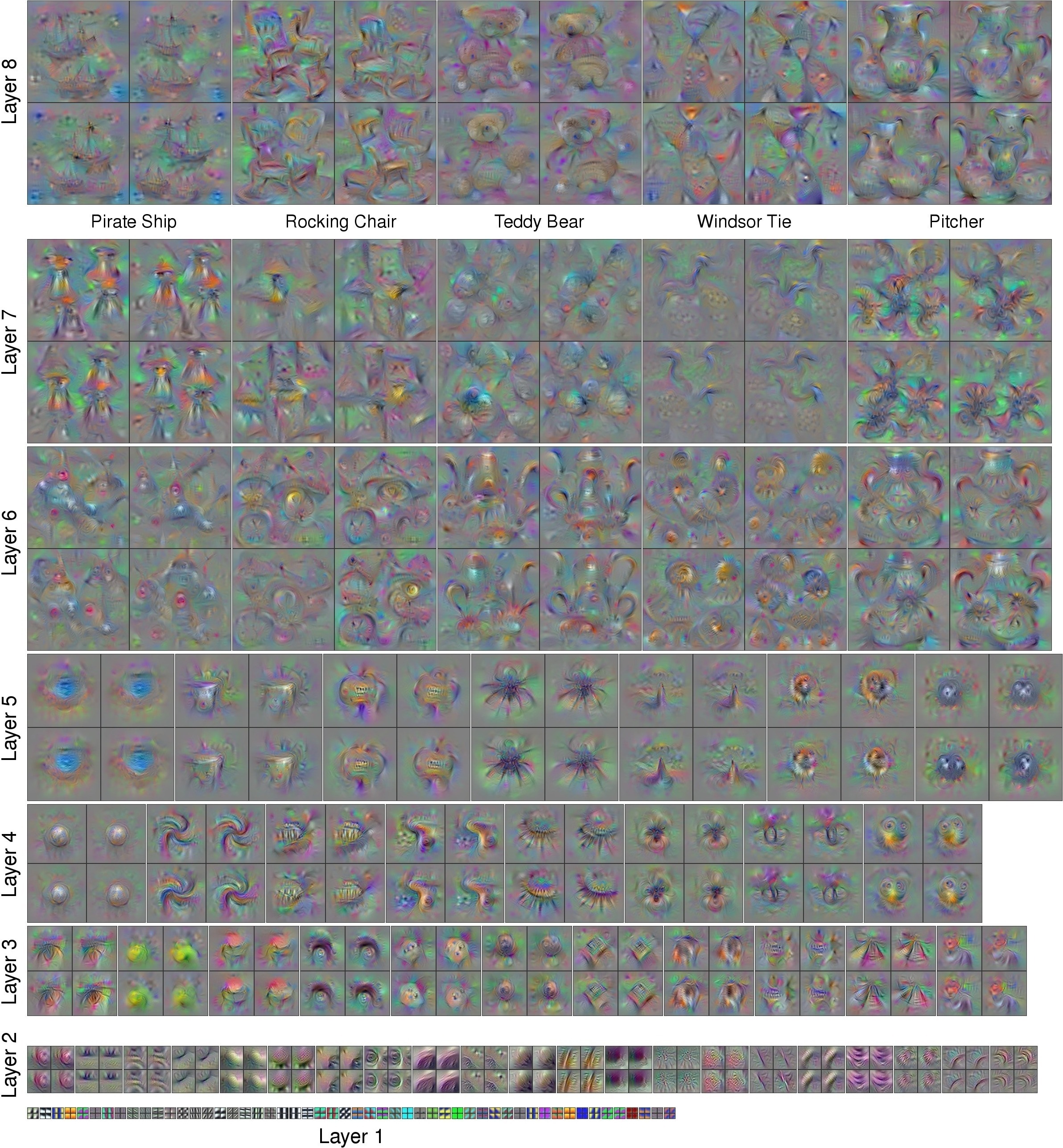}}
\caption{Visualization of example features of eight layers of a deep, convolutional neural network. The images reflect the true sizes of the features at different layers. In each layer, we show visualizations from 4 random gradient descent runs for each channel. While these images are hand picked to showcase the diversity and interpretability of the visualizations, one image for each filter of all five convolutional layers is shown in Figure S1 in supplementary information. One can recognize important features of objects at different scales, such as edges, corners, wheels, eyes, shoulders, faces, handles, bottles, etc. The visualizations show the increase in complexity and variation on higher layers, comprised of simpler components from lower layers. The variation of patterns increases with increasing layer number, indicating that increasingly invariant representations are learned. In particular, the jump from Layer 5 (the last convolution layer) to Layer 6 (the first fully-connected layer) brings about a large increase in variation. Best viewed electronically, zoomed in.
}
\figlabel{vis_all}
\end{center}
\vskip -0.2in
\end{figure*}

\section{Discussion and Conclusion}
\seclabel{conclusion}

We have introduced two visual tools for aiding in the interpretation
of trained neural nets.
Intuition gained from these tools may prompt ideas for improved methods and future research. Here we discuss several such ideas.

The interactive tool reveals that representations on later convolutional layers tend to be somewhat local, where channels correspond to specific, natural parts (e.g. wheels, faces) instead of being dimensions in a completely distributed code. That said, not all features correspond to natural parts, raising the possibility of a different decomposition of the world than humans might expect. These visualizations suggest that further study into the exact nature of learned representations --- whether they are local to a single channel or distributed across several --- is likely to be interesting (see Zhou et al.~\yrcite{zhou-2014-arXiv-object-detectors-emerge} for work in this direction). The locality of the representation also suggests that during transfer learning, when new models are trained atop the \layer{conv4} or \layer{conv5} representations, a bias toward sparse connectivity could be helpful because it may be necessary to combine only a few features from these layers to create important features at higher layers.

The second tool --- new regularizations that enable improved, interpretable, optimized visualizations of learned features --- will help researchers and practitioners understand, debug, and improve their models. The visualizations also reveal a new twist in an ongoing story. Previous studies
have shown that discriminative networks can easily be fooled or hacked by the addition of certain structured
noise in image space \cite{szegedy2013intriguing-properties-of-neural,nguyen-2014-arXiv-deep-neural-networks}.
An oft-cited reason for this property is that discriminative training leads networks
to ignore non-discriminative information in their input, e.g. learning to detect jaguars by matching the unique spots on their fur while ignoring the fact that they have four legs. For this reason it has been seen as a hopeless endeavor to create a generative model in which one randomly samples an $x$ from a broad distribution on the space of all possible images
and then iteratively transforms $x$ into a recognizable image by moving it to a region that satisfies both a prior $p(x)$ and posterior $p(y|x)$ for some class label $y$.
Past attempts have largely supported this view by producing unrealistic images using this method \cite{nguyen-2014-arXiv-deep-neural-networks,simonyan2013deep-inside-convolutional}.

However, the results presented here suggest an alternate possibility: the previously used priors may simply have been too weak (see \secref{high_freq} for one hypothesis of why a strong $p(x)$ model is needed). With the careful design or learning of a $p(x)$ model that biases toward realism,
one may be able to harness
the large number of parameters present in a discriminately learned $p(y|x)$ model
to generate realistic images by enforcing probability under both models simultaneously.
Even with the simple, hand-coded $p(x)$ models we use in this paper as regularizers, complex dependencies between distant pixels already arise (cf. the beetles with structure spanning over 100 pixels in \figref{vis_fc8}). This implies that the discriminative parameters also contain significant ``generative'' structure from the
training dataset; that is, the parameters encode
not only the jaguar's spots, but to some extent also its four legs.
With better, learned probabilistic models over the input and activations of higher layers, much more structure may be apparent. Work by Dai et al.~\yrcite{dai-2015-ICLR-generative-modeling-of-convolutional} shows some interesting results in this direction.
While the images generated in this paper are far from being photo-realistic, they do suggest that
transferring discriminatively trained parameters to generative models --- opposite the direction of the usual unsupervised pretraining approach --- may be a fruitful area for further investigation.


%

\section*{Acknowledgments} 

The authors would like to thank the NASA Space Technology Research Fellowship (JY) for funding, Wendy Shang, Yoshua Bengio, Brian Cheung, and Andrej Karpathy for helpful discussions, and Freckles the cat for her feline countenance.

{
  \small
\bibliography{bibdesk}

\begin{thebibliography}{23}
\providecommand{\natexlab}[1]{#1}
\providecommand{\url}[1]{\texttt{#1}}
\expandafter\ifx\csname urlstyle\endcsname\relax
  \providecommand{\doi}[1]{doi: #1}\else
  \providecommand{\doi}{doi: \begingroup \urlstyle{rm}\Url}\fi

\bibitem[Bergstra et~al.(2010)Bergstra, Breuleux, Bastien, Lamblin, Pascanu,
  Desjardins, Turian, Warde-Farley, and
  Bengio]{bergstra2010theano:-a-cpu-and-gpu-math-expression}
Bergstra, James, Breuleux, Olivier, Bastien, Fr{\'{e}}d{\'{e}}ric, Lamblin,
  Pascal, Pascanu, Razvan, Desjardins, Guillaume, Turian, Joseph, Warde-Farley,
  David, and Bengio, Yoshua.
\newblock Theano: a {CPU} and {GPU} math expression compiler.
\newblock In \emph{Proceedings of the Python for Scientific Computing
  Conference ({SciPy})}, June 2010.
\newblock Oral Presentation.

\bibitem[Collobert et~al.(2011)Collobert, Kavukcuoglu, and
  Farabet]{collobert-2011-torch7:-a-matlab-like-environment}
Collobert, Ronan, Kavukcuoglu, Koray, and Farabet, Cl{\'e}ment.
\newblock Torch7: A matlab-like environment for machine learning.
\newblock In \emph{BigLearn, NIPS Workshop}, number EPFL-CONF-192376, 2011.

\bibitem[Dai et~al.(2015)Dai, Lu, and
  Wu]{dai-2015-ICLR-generative-modeling-of-convolutional}
Dai, Jifeng, Lu, Yang, and Wu, Ying~Nian.
\newblock Generative modeling of convolutional neural networks.
\newblock In \emph{International Conference on Learning Representations
  (ICLR)}, 2015.

\bibitem[Deng et~al.(2009)Deng, Dong, Socher, Li, Li, and
  Fei-Fei]{deng2009imagenet:-a-large-scale-hierarchical}
Deng, Jia, Dong, Wei, Socher, Richard, Li, Li-Jia, Li, Kai, and Fei-Fei, Li.
\newblock Imagenet: A large-scale hierarchical image database.
\newblock In \emph{Computer Vision and Pattern Recognition, 2009. CVPR 2009.
  IEEE Conference on}, pp.\  248--255. IEEE, 2009.

\bibitem[Erhan et~al.(2009)Erhan, Bengio, Courville, and
  Vincent]{erhan2009visualizing-higher-layer-features}
Erhan, Dumitru, Bengio, Yoshua, Courville, Aaron, and Vincent, Pascal.
\newblock Visualizing higher-layer features of a deep network.
\newblock Technical report, Technical report, University of Montreal, 2009.

\bibitem[Glorot et~al.(2011)Glorot, Bordes, and
  Bengio]{glorot-2011-AISTATS-deep-sparse-rectifier}
Glorot, Xavier, Bordes, Antoine, and Bengio, Yoshua.
\newblock Deep sparse rectifier networks.
\newblock In \emph{Proceedings of the 14th International Conference on
  Artificial Intelligence and Statistics. JMLR W\&CP Volume}, volume~15, pp.\
  315--323, 2011.

\bibitem[Goodfellow et~al.(2013)Goodfellow, Warde-Farley, Lamblin, Dumoulin,
  Mirza, Pascanu, Bergstra, Bastien, and
  Bengio]{goodfellow2013pylearn2:-a-machine-learning-research}
Goodfellow, Ian~J, Warde-Farley, David, Lamblin, Pascal, Dumoulin, Vincent,
  Mirza, Mehdi, Pascanu, Razvan, Bergstra, James, Bastien, Fr{\'e}d{\'e}ric,
  and Bengio, Yoshua.
\newblock Pylearn2: a machine learning research library.
\newblock \emph{arXiv preprint arXiv:1308.4214}, 2013.

\bibitem[Goodfellow et~al.(2014)Goodfellow, Shlens, and
  Szegedy]{goodfellow-2014-arXiv-explaining-and-harnessing-adversarial}
Goodfellow, Ian~J, Shlens, Jonathon, and Szegedy, Christian.
\newblock {Explaining and Harnessing Adversarial Examples}.
\newblock \emph{ArXiv e-prints}, December 2014.

\bibitem[{Hannun} et~al.(2014){Hannun}, {Case}, {Casper}, {Catanzaro},
  {Diamos}, {Elsen}, {Prenger}, {Satheesh}, {Sengupta}, {Coates}, and
  {Ng}]{hannun-2014-arXiv-deep-speech:-scaling}
{Hannun}, A., {Case}, C., {Casper}, J., {Catanzaro}, B., {Diamos}, G., {Elsen},
  E., {Prenger}, R., {Satheesh}, S., {Sengupta}, S., {Coates}, A., and {Ng},
  A.~Y.
\newblock {Deep Speech: Scaling up end-to-end speech recognition}.
\newblock \emph{ArXiv e-prints}, December 2014.

\bibitem[Hinton et~al.(2012)Hinton, Srivastava, Krizhevsky, Sutskever, and
  Salakhutdinov]{hinton2012improving-neural-networks-by-preventing}
Hinton, Geoffrey~E, Srivastava, Nitish, Krizhevsky, Alex, Sutskever, Ilya, and
  Salakhutdinov, Ruslan~R.
\newblock Improving neural networks by preventing co-adaptation of feature
  detectors.
\newblock \emph{arXiv preprint arXiv:1207.0580}, 2012.

\bibitem[Jia et~al.(2014)Jia, Shelhamer, Donahue, Karayev, Long, Girshick,
  Guadarrama, and Darrell]{jia2014caffe:-convolutional-architecture}
Jia, Yangqing, Shelhamer, Evan, Donahue, Jeff, Karayev, Sergey, Long, Jonathan,
  Girshick, Ross, Guadarrama, Sergio, and Darrell, Trevor.
\newblock Caffe: Convolutional architecture for fast feature embedding.
\newblock \emph{arXiv preprint arXiv:1408.5093}, 2014.

\bibitem[Krizhevsky et~al.(2012)Krizhevsky, Sutskever, and
  Hinton]{krizhevsky2012imagenet-classification-with-deep}
Krizhevsky, Alex, Sutskever, Ilya, and Hinton, Geoff.
\newblock Imagenet classification with deep convolutional neural networks.
\newblock In \emph{Advances in Neural Information Processing Systems 25}, pp.\
  1106--1114, 2012.

\bibitem[Lin et~al.(2014)Lin, Maire, Belongie, Hays, Perona, Ramanan,
  Doll{\'{a}}r, and Zitnick]{lin-2014-arXiv-microsoft-coco-common}
Lin, Tsung{-}Yi, Maire, Michael, Belongie, Serge, Hays, James, Perona, Pietro,
  Ramanan, Deva, Doll{\'{a}}r, Piotr, and Zitnick, C.~Lawrence.
\newblock Microsoft {COCO:} common objects in context.
\newblock \emph{CoRR}, abs/1405.0312, 2014.
\newblock URL \url{http://arxiv.org/abs/1405.0312}.

\bibitem[{Mahendran} \& {Vedaldi}(2014){Mahendran} and
  {Vedaldi}]{mahendran-2014-arXiv-understanding-deep-image}
{Mahendran}, A. and {Vedaldi}, A.
\newblock {Understanding Deep Image Representations by Inverting Them}.
\newblock \emph{ArXiv e-prints}, November 2014.

\bibitem[{Nguyen} et~al.(2014){Nguyen}, Yosinski, and
  Clune]{nguyen-2014-arXiv-deep-neural-networks}
{Nguyen}, Anh, Yosinski, Jason, and Clune, Jeff.
\newblock {Deep Neural Networks are Easily Fooled: High Confidence Predictions
  for Unrecognizable Images}.
\newblock \emph{ArXiv e-prints}, December 2014.

\bibitem[{Schroff} et~al.(2015){Schroff}, {Kalenichenko}, and
  {Philbin}]{schroff-2015-arXiv-facenet:-a-unified-embedding}
{Schroff}, F., {Kalenichenko}, D., and {Philbin}, J.
\newblock {FaceNet: A Unified Embedding for Face Recognition and Clustering}.
\newblock \emph{ArXiv e-prints}, March 2015.

\bibitem[Simonyan et~al.(2013)Simonyan, Vedaldi, and
  Zisserman]{simonyan2013deep-inside-convolutional}
Simonyan, Karen, Vedaldi, Andrea, and Zisserman, Andrew.
\newblock Deep inside convolutional networks: Visualising image classification
  models and saliency maps.
\newblock \emph{arXiv preprint arXiv:1312.6034, presented at ICLR Workshop
  2014}, 2013.

\bibitem[Szegedy et~al.(2013)Szegedy, Zaremba, Sutskever, Bruna, Erhan,
  Goodfellow, and Fergus]{szegedy2013intriguing-properties-of-neural}
Szegedy, Christian, Zaremba, Wojciech, Sutskever, Ilya, Bruna, Joan, Erhan,
  Dumitru, Goodfellow, Ian~J., and Fergus, Rob.
\newblock Intriguing properties of neural networks.
\newblock \emph{CoRR}, abs/1312.6199, 2013.

\bibitem[Taigman et~al.(2014)Taigman, Yang, Ranzato, and
  Wolf]{taigman-2014-CVPR-deepface:-closing-the-gap-to-human-level}
Taigman, Yaniv, Yang, Ming, Ranzato, Marc'Aurelio, and Wolf, Lior.
\newblock Deepface: Closing the gap to human-level performance in face
  verification.
\newblock In \emph{Computer Vision and Pattern Recognition (CVPR), 2014 IEEE
  Conference on}, pp.\  1701--1708. IEEE, 2014.

\bibitem[Torralba \& Oliva(2003)Torralba and
  Oliva]{torralba-2003-Network-statistics-of-natural-image}
Torralba, Antonio and Oliva, Aude.
\newblock Statistics of natural image categories.
\newblock \emph{Network: computation in neural systems}, 14\penalty0
  (3):\penalty0 391--412, 2003.

\bibitem[{Yosinski} et~al.(2014){Yosinski}, {Clune}, {Bengio}, and
  {Lipson}]{yosinski-2014-NIPS-how-transferable-are-features-in-deep}
{Yosinski}, J., {Clune}, J., {Bengio}, Y., and {Lipson}, H.
\newblock {How transferable are features in deep neural networks?}
\newblock In Ghahramani, Z., Welling, M., Cortes, C., Lawrence, N.D., and
  Weinberger, K.Q. (eds.), \emph{Advances in Neural Information Processing
  Systems 27}, pp.\  3320--3328. Curran Associates, Inc., December 2014.

\bibitem[Zeiler \& Fergus(2013)Zeiler and
  Fergus]{zeiler2013visualizing-and-understanding-convolutional}
Zeiler, Matthew~D and Fergus, Rob.
\newblock Visualizing and understanding convolutional neural networks.
\newblock \emph{arXiv preprint arXiv:1311.2901}, 2013.

\bibitem[Zhou et~al.(2014)Zhou, Khosla, Lapedriza, Oliva, and
  Torralba]{zhou-2014-arXiv-object-detectors-emerge}
Zhou, Bolei, Khosla, Aditya, Lapedriza, {\`{A}}gata, Oliva, Aude, and Torralba,
  Antonio.
\newblock Object detectors emerge in deep scene cnns.
\newblock \emph{CoRR}, abs/1412.6856, 2014.
\newblock URL \url{http://arxiv.org/abs/1412.6856}.

\end{thebibliography}
\bibliographystyle{icml2015}
}

\clearpage

\renewcommand{\thesection}{S\arabic{section}}
\renewcommand{\thesubsection}{\thesection.\arabic{subsection}}

\newcommand{\beginsupplementary}{%
        \setcounter{table}{0}
        \renewcommand{\thetable}{S\arabic{table}}%
        \setcounter{figure}{0}
        \renewcommand{\thefigure}{S\arabic{figure}}%
        \setcounter{section}{0}
     }

\beginsupplementary


\icmltitlerunning{\suptitlrunning}

\twocolumn[
  \icmltitle{\suptitl}

\icmlauthor{Jason Yosinski}{yosinski@cs.cornell.edu}
\icmlauthor{Jeff Clune}{jeffclune@uwyo.edu}
\icmlauthor{Anh Nguyen}{anguyen8@uwyo.edu}
\icmlauthor{Thomas Fuchs}{fuchs@caltech.edu}
\icmlauthor{Hod Lipson}{hod.lipson@cornell.edu}

\icmlkeywords{convolutional neural networks, neural networks, visualization}

\vskip 0.3in
]


\section{Why are gradient optimized images dominated by high frequencies?}
\seclabel{high_freq}

In the main text we mentioned that images produced by gradient ascent to maximize the activations of neurons in convolutional networks tend to be dominated by high frequency information (cf.  the left column of \figref{regularization_sweep}). One hypothesis for why this occurs centers around the differing statistics of the activations of channels in a convnet. The \layer{conv1} layer consists of blobs of color and oriented Gabor edge filters of varying frequencies. The average activation values (after the rectifier) of the edge filters vary across filters, with low frequency filters generally having much higher average activation values than high frequency filters. In one experiment we observed that the average activation values of the five lowest frequency edge filters was 90 versus an average for the five highest frequency filters of 5.4, a difference of a factor of 17 (manuscript in preparation)\footnote{Li, Yosinski, Clune, Song, Hopcroft, Lipson. 2015. How similar are features learned by different deep neural networks? In preparation.}\textsuperscript{,}\footnote{Activation values are averaged over the ImageNet validation set, over all spatial positions, over the channels with the five \{highest, lowest\} frequencies, and over four separately trained networks.}. The activation values for blobs of color generally fall in the middle of the range.
This phenomenon likely arises for reasons related to the $1/f$ power spectrum of natural images in which low spatial frequencies tend to contain higher energy than high spatial frequencies \citep{torralba-2003-Network-statistics-of-natural-image}.



Now consider the connections from the \layer{conv1} filters to a single unit on \layer{conv2}.
In order to merge information from both low frequency and high frequency \layer{conv1} filters, the connection weights from high frequency \layer{conv1} units may generally have to be larger than connections from low frequency \layer{conv1} units in order to allow both signals to affect the \layer{conv2} unit's activation similarly.
If this is the case, then due to the larger multipliers, the activation of this particular \layer{conv2} unit is affected more by small changes in the activations of high frequency filters than low frequency filters. Seen in the other direction: when gradient information is passed from higher layers to lower layers during backprop, the partial derivative arriving at this \layer{conv2} unit (a scalar) will be passed backward and multiplied by larger values when destined for high frequency \layer{conv1} filters than low frequency filters. Thus, following the gradient in pixel space may tend to produce an overabundance of high frequency changes instead of low frequency changes.

The above discussion focuses on the differing statistics of edge filters in \layer{conv1}, but note that activation statistics on subsequent layers also vary across each layer.\footnote{We have observed that statistics vary on higher layers, but in a different manner: most channels  on these layers have similar average activations, with most of the variance across channels being dominated by a small number of channels with unusually small or unusually large averages (Li, Yosinski, Clune, Song, Hopcroft, Lipson. 2015. How similar are features learned by different deep neural networks? In preparation.)} This may produce a similar (though more subtle to observe) effect in which rare higher layer features are also overrepresented compared to more common higher layer features.

Of course, this hypothesis is only one tentative explanation for why high frequency information dominates the gradient. It relies on the assumption that the average activation of a unit is a representative statistic of the whole distribution of activations for that unit. In our observation this has been the case, with most units having similar, albeit scaled, distributions. However, more study is needed before a definitive conclusion can be reached.

\section{Conv Layer Montages}

One example optimized image using the hyperparameter settings from the third row of \tabref{paramTable} for every filter of all five convolutional layers is shown in \figref{layer_montages}.

\begin{figure*}[ht]
\vskip 0.2in
\begin{center}
\centerline{\includegraphics[width=1\textwidth]{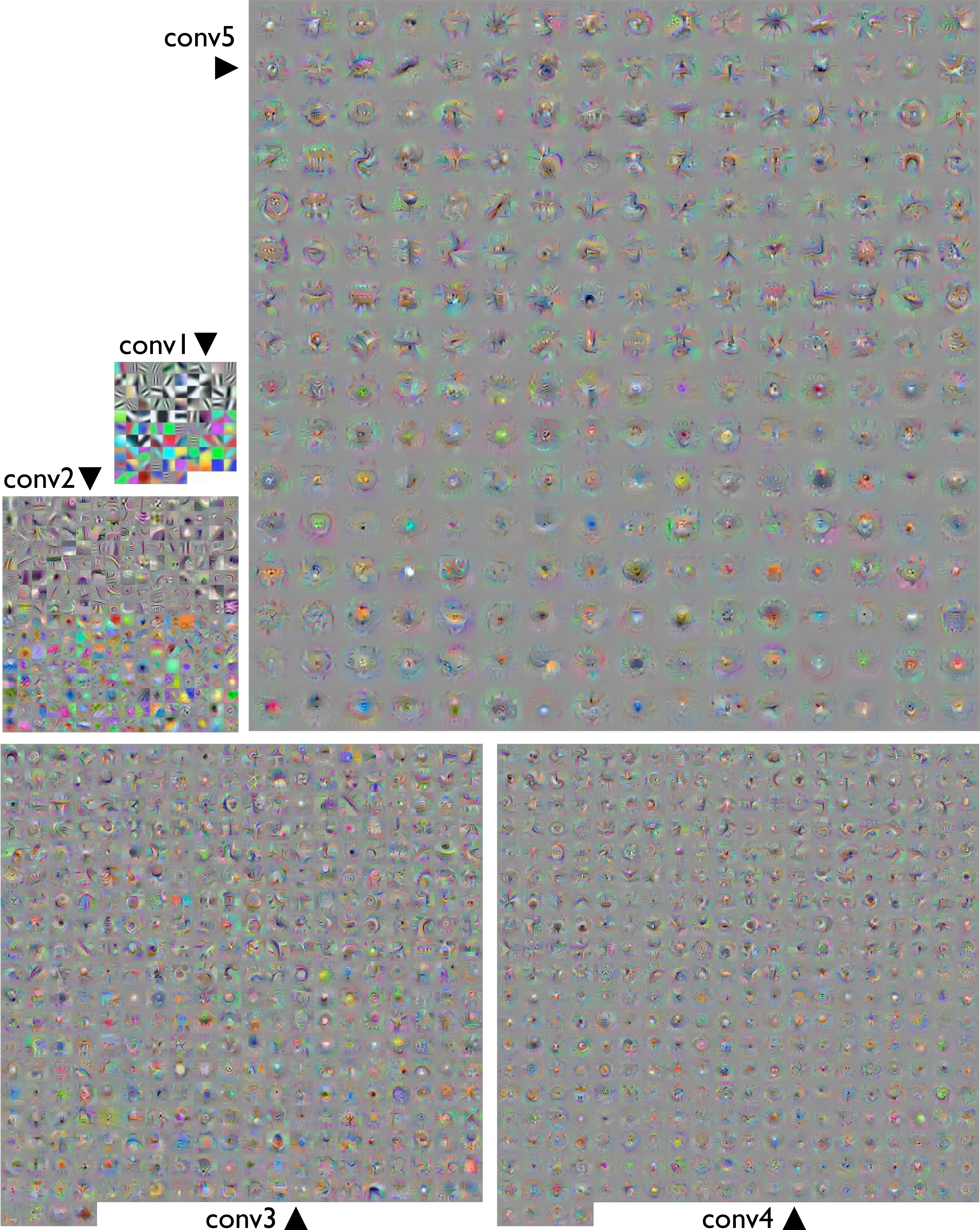}}
\caption{One optimized, preferred image for every channel of all five convolutional layers. These images were produced with the hyperparameter combinations from the third row of \tabref{paramTable}. Best viewed electronically, zoomed in.}
\figlabel{layer_montages}
\end{center}
\vskip -0.2in
\end{figure*}

\end{document}